# Title

Full: Reimagining partial thickness keratoplasty: An eye mountable robot for autonomous big bubble needle insertion

Short: Autonomous needle insertion for DALK

## Authors

Y. Wang [1,*,**], J. D. Opfermann [2,*], J. Yu [1], H. Yi [1], J. Kaluna [2], R. Biswas [2], R. Zuo [1], W. Gensheimer [3,4], A. Krieger [2], J. U. Kang [1,**]

## Affiliations

[1] Department of Electrical and Computer Engineering, Johns Hopkins University, Baltimore, MD 21211, USA.

[2] Department of Mechanical Engineering, Johns Hopkins University, Baltimore, MD 21211, USA.

[3] Department of Ophthalmology, White River Junction VA Medical Center, White River Junction, VT 05001, USA.

[4] Department of Ophthalmology, Dartmouth-Hitchcock Medical Center, Lebanon, NH 03766, USA.

* Co-First Authors.
** Co-Corresponding Authors: Jin U. Kang (jkang@jhu.edu), Yaning Wang (ywang511@jh.edu).

## Abstract

Autonomous surgical robots have demonstrated significant potential to standardize surgical outcomes, driving innovations that enhance safety and consistency regardless of individual surgeons' experience. Deep anterior lamellar keratoplasty (DALK), a partial thickness corneal transplant surgery aimed at replacing the anterior part of cornea above Descemet's membrane (DM), would greatly benefit from an autonomous surgical approach as it highly relies on surgeon skill with high perforation rates. In this study, we proposed a novel autonomous surgical robotic system (AUTO-DALK) based on a customized neural network capable of precise needle control and consistent big bubble demarcation on cadaver and live rabbit models. We demonstrate the feasibility of an AI-based image-guided vertical drilling approach for big bubble generation, in contrast to the conventional horizontal needle approach. Our system integrates an optical coherence tomography (OCT) fiber optic distal sensor into the eye-mountable micro robotic system, which automatically segments OCT M-mode depth signals to identify corneal layers using a custom deep learning algorithm. It enables the robot to autonomously guide the needle to targeted tissue layers via a depth-controlled feedback loop. We compared autonomous needle insertion performance and resulting pneumodissection using AUTO-DALK against 1) freehand insertion, 2) OCT sensor guided manual insertion, and 3) teleoperated robotic insertion, reporting significant improvements in insertion depth, pneumodissection depth, task completion time, and big bubble formation. *Ex vivo* and *in vivo* results indicate that the AI-driven, AUTO-DALK system, is a promising solution to standardize pneumodissection outcomes for partial thickness keratoplasty.

## Summary

We demonstrate an AI-based autonomous robotic system for needle insertion during the big bubble DALK procedure.

# INTRODUCTION

Autonomous surgical robots are expected to become an integral part of future surgeries. Their promise to standardize surgical outcomes, ensure patient safety, and democratize patient care has led to technical and clinical innovations that revolutionized healthcare [Dupont-2021-retro, Andras-2020-artificial]. Autonomous robotic systems can eliminate unconscious bias in a surgeon's performance due to differences in surgical training, education, and technique [Xu-2016-surgeon], surgical center [Pasquali-2020-variation], and will reduce a surgeon's learning curve [Kassite-2019-system]. Historically, autonomous surgical robots were limited to boney tissue surgeries such with the Mako Surgical System (Stryker, Kalamazoo, MI) [Roche-2021-mako], Tsolution One (THINK Surgical, Fremont, CA) [Liow-2017-think], and the ExcelsiusGPS (Globus Medical, Audubon, PA) [Shi-2024-spine], with soft tissue applications limited to contactless therapies such as CyberKnife (Accuray, Madison, WI) tumor ablation [Kilby-2010-knife]. However, advances in artificial intelligence and machine learning methods have increased the utility of robotic system for more advanced surgical techniques which now include needle pick-up and grasping [Kim-2024-transformer], surgical cutting of phantom tissues [Murali-2015-observation], and feature extraction in surgical planning [Lu-2021-super]. In one landmark study, neural networks were used to overcome limitations during intraoperative tissue tracking to perform the first autonomous laparoscopic intestinal anastomosis in a living model [Saeidi-2022-autonomous].

Autonomous surgical robots can be especially beneficial for microsurgeries, where surgical outcomes rely on precise and highly dexterous motions in soft tissue, and variations due to muscle tremor or other uncontrollable human factors that affect surgeons' day-to-day performance [MacDonald-2005-learning]. Eye surgeries for instance, often require micron scale motions which are challenging for even the most experienced surgeon. Teleoperated systems for intraocular manipulation [Nasseri-2013-surgery] and cannula insertion [Zhao-2023-cannula] were proposed to scale surgical motions to the micron scale, while hand-held robotic tools [cheon-2017-motorized, lee-2021-cnn, huang-2012-motion] and a steady-hand robotic system were developed to remove surgical tremor for hand guided microsurgery [Taylor-1999-steady, Mitchell-2007-steady]. However, clinical autonomous systems are limited to contactless therapies that minimize tissue deformation such as LASIK with VisuMax (Zeiss, Oberkochen, Germany) [Blum-2009-lasik].

Big bubble Deep Anterior Lamellar Keratoplasty (DALK) is a partial thickness corneal transplant microsurgery that would be particularly useful to automate. In a big bubble DALK procedure, an ophthalmologist inserts a needle into the patient's eye up to 90% depth of the total stromal thickness. Air is then injected to pneumodissect the Descement's membrane from the stromal tissue, so that the host cornea can be removed with a trephine while leaving the endothelial layer intact (Fig. 1a). Preserving the endothelial layer reduces the risk of donor tissue rejection and intraoperative complications compared to full thickness corneal transplants [Sugita-1997-dalk, Cheng-2013-comparison]. Despite clear patient benefits, the DALK technique is seldom used because needle insertion is technically challenging. A healthy human cornea is 515 μm thick which means the needle must be inserted to a depth just 51.5 μm away from perforating the Descement's membrane. Achieving the necessary micron-level depth accuracy for successful stromal pneumodissection is challenging for even the most experienced surgeon, and the reported rates of failed pneumodissection and intraocular perforation are 15% and 11.7%, respectively. These results are even higher for less trained surgeons [Busin-2016-outcomes, Reinhart-2011-report, Unal-2010-conversion]. If the Descemet's membrane is perforated, the procedure must be converted to a full thickness transplant which denies the patient access to better surgical outcomes. Perhaps most frustrating, even when the needle is inserted to the correct depth, a horizontal freehand approach may still generate incomplete and unpredictable pneumo-dissection (Fig. 1b) [Opfermann-2024-vertical]. Incomplete pneumodissection of the stroma requires the surgeons to manually dissect the stroma from Dua's layer, adding time and variability to the surgical outcomes. An autonomous robotic system for big bubble DALK procedures is an attractive solution to simplify the critical step of needle insertion so that any surgeon, independent of training and skill, can achieve the ideal surgical outcomes associated with a partial thickness transplant technique.

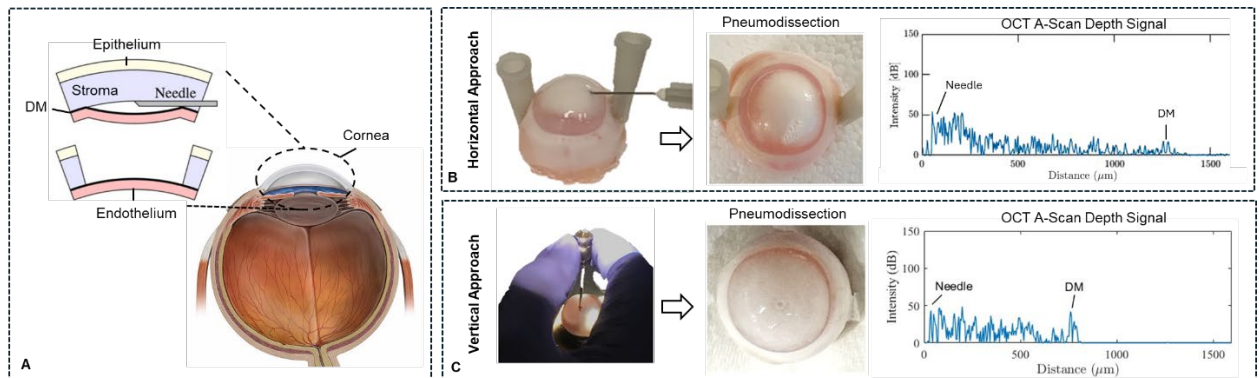

**Figure 1. Deep Anterior Lamellar Keratoplasty (DALK) surgical approach.** (A) A diagram illustrating the critical step if inserting a pneumodissection needle to the level of Descemet's membrane to perform endothelium sparing pneumodissection. (B) Surgical approach, representative pneumodissection, typical OCT-Ascan depth signal for a traditional horizontal needle approach (B) and vertical needle approach (C).

State of the art robots for autonomous DALK needle insertion aim to replicate the clinical approach of inserting the pneumodissection needle along an angled or horizontal trajectory, preserving the viewing of the needle from a surgical microscope [Draelos-2020-oct]. These systems use intraoperative OCT from the surgical microscope to generate depth profiles of the cornea so that the needle can be guided along a horizontal trajectory for pneumodissection of cadaver eyes. It is unknown how these systems will handle sudden and unexpected movements of the eye in living tissue studies which is still significant under deep anesthesia [Laborit-1977-movement, Zuo-2022-regression, Nordlund-2003-cataract]. Large corneal deformations are expected during needle insertion [Edwards-2022-data, Draelos-2020-oct], and accurately modeling the deformation for motion planning presents a challenge. Learning from Demonstration (LfD) and reinforcement learning have been used to model the needle-tissue interaction [Keller-2020-robotic], and real time OCT image segmentation and correction algorithms have been proposed to track the needle tip and endothelium [Park-2020-deep]. Eye mountable robots overcome these limitations by physically attaching the robot to the eye, which maintains needle alignment during involuntary motions [Opfermann-2024-design]. Image guidance is performed using intraoperative A-scan OCT acquired from an imaging fiber embedded in the pneumodissection needle, which can degrade in signal quality with a traditional horizonal approach (Fig. 1b). An eye mountable robot with a vertical needle approach would be more desirable for autonomous needle insertion as 1) vertical needle injection pressures are more uniform which generates mode consistent pneumodissection than the traditional horizontal approach, even with freehand insertion [Opfermann-2024-vertical], and 2) the imaging quality of a downward facing OCT fiber probe is significantly improved which enables autonomous segmentation of the Descement's membrane and depth-controlled feedback for robotic needle positioning (Fig. 1c). To this end, our team aspires to reimagine the critical step of needle insertion during the DALK procedure by employing a robotic vertical needle approach for big bubble pneumodissection, something that has yet to be shown clinically feasible. We aim to demonstrate autonomous OCT guided vertical needle insertion is both safe and effective, resulting in significantly more accurate needle placement and consistent pneumodissection of the deep stroma for the DALK procedure.

As our first contribution, we present a novel surgical robot called AUTO-DALK (Autonomous DALK) (Fig. 2) that redefines the surgical paradigm for partial thickness corneal transplant surgeries and reimagines the procedure as would be idealized for autonomous robotic control. Instead of a horizontal needle approach to big bubble generation, we propose an AI-driven image guided vertical approach for more consistent deep stromal pneumodissection. The vertical needle robot combines light weight motors with a high-resolution optical coherence tomography (OCT) depth sensing needle to enable micron scale needle positioning for big bubble DALK procedures. OCT depth signals are autonomously segmented using a customized topology-preserving shape-regularized U-Net (TS-U-Net) architecture to identify and track the epithelium, Descemet's membrane, and target needle depth tissue layers. Needle positioning in the deep stroma is performed autonomously using a depth-controlled feedback loop. The system is light weight, with minimal footprint, and features a vacuum channel system that facilitates direct attachment to the patient's eye to ensure there is proper alignment during the procedure even in the instance of involuntary eye movements. The proposed high resolution microneedle needle and OCT based imaging system may also be impactful in other surgical areas such as intraoperative guidance of needles for brain and prostate tumor biopsy or ablation.

The second contribution of this work is demonstrating significantly improved pneumodissection of the deep stroma as compared to an expert surgeon trained in the DALK procedure. We conduct an extensive *ex vivo* study to evaluate the robotic system to perform deep stromal needle insertion and resulting pneumodissection as compared to the clinical horizontal free hand approach, a freehand vertical approach using the vertical needle with embedded OCT fiber-optic depth sensor, a teleoperated vertical robotic approach, and our proposed autonomous vertical robotic approach. Additionally, we evaluate the system using an *in vivo* rabbit model to compare pneumodissection performance between teleoperated and autonomous modes of robotic operation to demonstrate clinical feasibility of the system. Two AI based networks are used for robot guidance to demonstrate how geometric constraints and variability in *in vivo* data can be used to improve the machine learning model for the *in vivo* approach. We conclude with statistical improvements in key clinical metrics to justify our novel vertical approach as being both safe and effective for potential clinical translation in future work.

**RESULTS**

In this section, we describe the design and clinical workflow of the AUTO-DALK system for autonomous needle insertion, followed by *in vivo* evaluation in a live rabbit model, a comprehensive *ex vivo* analysis of the system compared to expert surgeons, and include a robust evaluation of the AI-based corneal tissue segmentation and tracker.

**System Design and Workflow for Autonomous Big Bubble Needle Insertion**

The AUTO-DALK system, featuring an automatic control strategy, is illustrated in Fig. 2. The robotic system employs a vertically inserted razor edge needle approach to penetrate the cornea to the desired depth. The robot comprises two symmetric micro-stepper motors (RobotDigg, Shanghai, China), a custom made common-path swept-source OCT (CP-SS-OCT) fiber sensor integrated needle, and a mirror (Fig. 2a). Using an optical fiber embedded in the vertical needle, the OCT system records M-scan depth information during the procedure which is displayed to the user so that variations in needle depth can are visible in real time. The temporal depth information is then passed to a deep learning algorithm, that autonomously segments key tissue layers of the cornea such as the epithelium, Descemet's membrane, and target needle depth. The segmentation is performed real time so that the depth information of all key tissue layers is accurately tracked during the procedure, enabling real time depth-controlled feedback of the needle. All OCT depth information is provided to the surgeon through a custom graphical user interface (GUI) (Fig. 2b). The system has two modes of operation 1) *Teleoperation Mode* whereby the surgeon is able to manually advance the robot based on a user defined step size, and 2) *Autonomous Mode* whereby the surgeon can set a target depth relative to the Descement's membrane, and the robot autonomously segments, interprets, and advances the needle to the desired tissue depth. The robotic system incorporates a right-angle mirror so that an operating surgeon can directly view insertion of the needle in the cornea using a standard surgical microscope as in Fig. 2a.

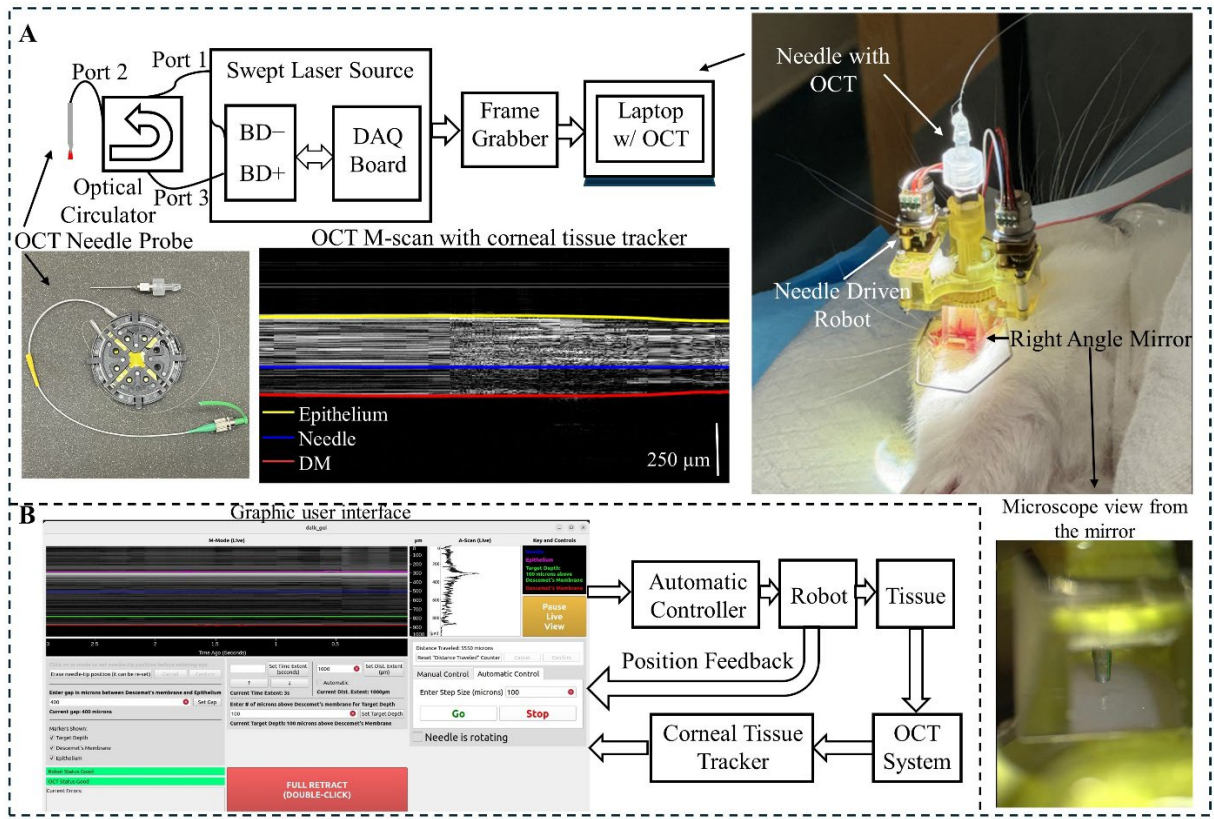

**Figure 2. Autonomous big bubble DALK procedure.** (A) The components of the AUTO-DALK system including needle drive platform with embedded mirror, and CP-SS-OCT system. (B) Control architecture of the autonomous control strategy for AUTO-DALK robot.

To operate the robot in *Autonomous Mode*, the distance between the needle tip and embedded fiber sensor tip (as measured during probe manufacturing) is inputted into the GUI. The AUTO-DALK system converts this value to an optical distance based on the corneal tissue refractive index correction (see "Imaging System" section for details). The needle is retracted to a zero position so that the AUTO-DALK system can be secured to the apex of the cornea using the vacuum syringe, without the needle perforating the cornea. The operator will then manually advance the needle via the GUI in manual mode until the needle tip contacts the epithelium, as viewed through the right-angle mirror and surgical microscope, and the user will also specify the insertion step size and target needle depth above the Descemet's Membrane for the final needle position. Once the *Autonomous Mode* is initiated, the AUTO-DALK system advances the needle to the desire target depth. The system continuously monitors the distance between the current and target needle depths using the AI-based tissue segmentation and tracking algorithm. Based on real-time feedback, the system dynamically adjusts the motor step size, as outlined in section *Materials and Methods - Autonomous Control Strategy*. When the needle reaches the target depth, or the mechanical travel limit, the system stops advancing the needle and notifies the operator that the needle depth had been reached. The operating surgeon selects a retraction routine which raises the vertical needle out of the cornea. Following this, the vacuum attachment is released by depressing the vacuum syringe, and air is manually injected by the surgeon via a syringe for pneumodissection.

## *In vivo* Big Bubble DALK procedure

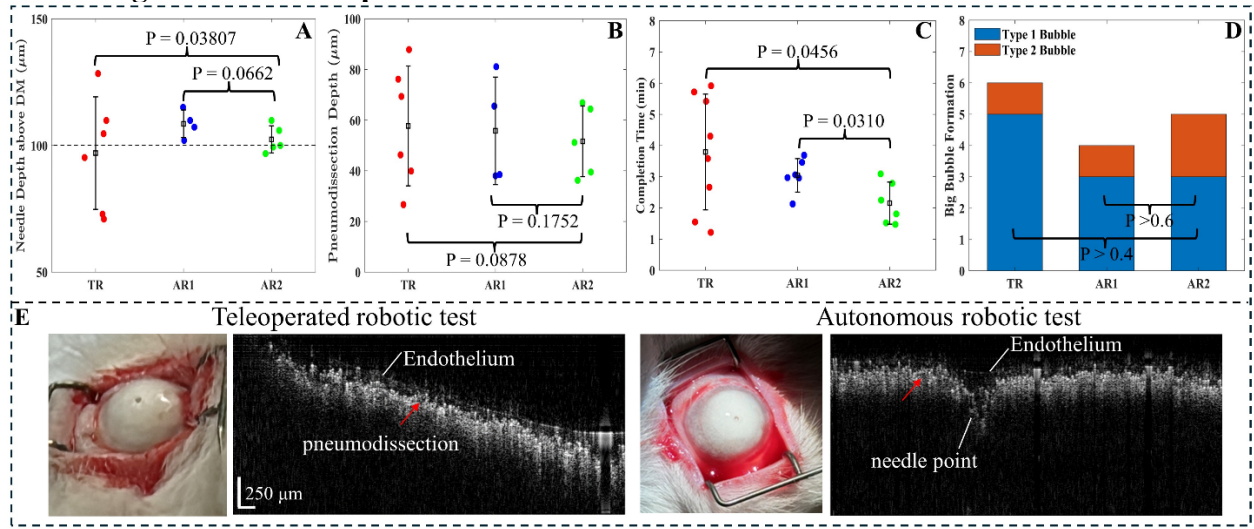

**Figure 3. The results of *ex vivo* big bubble test via TR (n = 8), AR1 (n = 5) and AR2 (n = 6).** (A) needle depth above Descemet's membrane, (B) pneumodissection depth, (C) task completion time, (D) big bubble formation, (E) representative pneumodissection results of color images with blanching of the corneal stroma and OCT images (endothelial side up) via TR, and AR2.

We conducted an *in vivo* live rabbit experiment to evaluate the performance of the AUTO-DALK system to generate consistent pneumodissection of the deep stroma when controlled using either the *Teleoperated* or *Autonomous* Mode (protocol 00002276) at Dartmouth-Hitchcock Medical Center. The study used a white albino live rabbit model weighing ~ 3.3 kg, 3 to 5 months old from Charles River Laboratories. A rabbit eye model was selected for its prevalence as a surgical training model in ophthalmic surgery. The study groups included two autonomous robotic tests via AUTO-DALK (AR1, n = 5 eyes; AR2, n = 6 eyes), and one teleoperated robotic test via AUTO-DALK (TR, n = 8 eyes). The difference between the study groups AR1 and AR2 is the deep learning model used to autonomously segment and track the *in vivo* cornea tissue layers. In AR1, the deep learning model was trained using 62 samples from *ex vivo* and 8 samples from *in vivo* eyes. After conducting AR1, we optimized our deep learning model by incorporating a shape regularization method (described further in *Results - TS-U-Net based tissue tracking*) and by adding new *in vivo* datasets collected during the AR1 study that included physiological features such as the iris. The new segmentation and training model was used for the subsequent autonomous robotic test, AR2. The robot in AR2 also incorporated a two-piece needle guide mechanism that reduced needle runout during the procedure as described in the *Discussion*.

The AUTO-DALK robot was first attached to the apex of the cornea using the vacuum channel as in Fig. 2a (and supplemental Figure S1). For *Teleoperation Mode* experiments, the surgeon was instructed to insert the needle to a similar depth as would be done for a clinical case. During *Autonomous Mode* robotic experiments, the needle was driven to the target position using a feedback control loop with the OCT depth sensing and AI-based tissue tracking algorithm. The average central thickness of the *in vivo* corneas was measured intraoperatively as 369.4 μm, with a standard deviation of 24.2 μm. Compared with aged, cadaver corneal tissues, the live corneal tissue has lower stiffness, and it is more flexible [Shimmura-2006-deep]. Thus, the target insertion depth above the Descemet's Membrane was set to 100 μm (~72.9% of the total thickness), a conservative limit so that the robot would adhere to the animal care protocols and ensure procedural safety during these initial *in vivo* experiments. After the vertical needle was inserted in all cases, AUTO-DALK was removed from the eye and a blunt tipped needle with a 5 ml syringe inserted into the needle track and use to administer pneumodissection of the deep stroma.

The evaluation metrics for surgical outcomes included the perforation rate, needle depth above Descemet's membrane, pneumodissection depth, task completion time, and classification of big bubbles. The needle depth above Descemet's membrane was calculated as the distance between Descemet's membrane and the needle tip at the final position based on OCT signals. Pneumodissection depth was calculated as an average distance between the endothelium and the partial bubble according to the postoperative OCT B-mode images. The task completion time was measured from the moment the needle touched the epithelium layer until it reached the target position. Big bubbles were classified based on the color, shape and corresponding OCT images. All the measurements were reviewed by an experienced surgeon.

The average needle depth, along with standard deviation, were computed to be $97.0 \pm 20.3$ μm, $108.6 \pm 4.7$ μm, and $102.4 \pm 4.8$ μm for TR, AR1, and AR2, respectively (Fig 3a). The perforation rate was measured as 25.0%, 20.0%, and 16.7% for TR, AR1, and AR2, respectively. The AUTO-DALK system demonstrated more consistent needle depths, and shallower needle depths in the autonomous mode compared with the teleoperation mode ($P = 0.03807$), resulting in a lower perforation rate. Higher insertion depths (44.6 μm, 88.6% of the total thickness) were more likely to cause perforation of Descemet's membrane, and perforation could occur from either the needle insertion or air injection. As a result, further evaluations on the function of needle depth and perforation rate due to pneumodissection of a vertical needle are necessary. The average pneumodissection depth (Fig. 3b) for AR2 was measured at $51.6 \pm 12.5$ μm, showing marginally significant improvement in consistency compared with TR ($P=0.0878$). No statistically significant differences in pneumodissection depth were observed between AR1 and AR2 ($P = 0.1752$). The AUTO-DALK system completed the needle insertion task faster when operating autonomously (AR1, AR2), with average completion times of 182.7 s and 128.2 s, respectively, compared with TR ($P < 0.05$) (Fig. 3c). No significant differences in big bubble formation (as measured by the percentage of formed type 1 bubbles) were noted between three groups ($P > 0.3$) (Fig. 3d). Compared with AR1, AR2 achieved a final needle depth marginally closer to the ideal target depth ($P = 0.0662$) with shorter completion time ($P = 0.0310$), attributed to the performance of the improved deep learning model. The evaluation result also indicates the proposed AUTO-DALK system has good robustness and consistency in separating the Descemet's membrane from the stroma, with deeper pneumodissection depth compared with *ex vivo* studies ($P = 0.0370$). The perforation in the AR1 group occurred because the iris was misclassified as the endothelium, preventing the robot from identifying the ideal target depth. Subsequently, we optimized the AI algorithm to accurately distinguish the Descemet's membrane from surrounding tissues. The perforation in the AR2 group resulted from a large step size. There was one eye in the AR2 group where autonomous needle insertion was attempted but could not be initiated due to a typographical error entered when executing the AI algorithm. This error resulted in unstable data transmission, leading to misplaced OCT M-scans and segmented results. In another sample, poor signal quality from the OCT fiber in the needle resulted in system from advancing the needle before the cornea was penetrated. The trial was successful after replacing the needle and OCT fiber sensor.

*Ex vivo* **big bubble DALK procedure**

We conducted a comprehensive *ex vivo* study to evaluate our Autonomous robotic DALK system against the clinical standard of care, to perform deep stromal pneumodissection using the big bubble technique (See supplemental Figure S2 for illustrations of the test setups). For these experiments, 24 five-day-old rabbit eyes (Pel-Freez Biologicals, Rogers, AR) from albino rabbits (4.75-5.75 lbs., approximately 8-12 weeks old, mixed gender) were used. For all tests, eyes were held in place by suction of a 3D printed eye mount. Using a tonometer, the intraocular pressure was measured to be within the normal physiological range prior to each experiment. Four different approaches were conducted: vertical freehand insertion (FH), vertical OCT sensor guided insertion (SG), teleoperated robotic insertion (TR), and autonomous robotic insertion (AR). To minimize bias and the effect of learning curve, Excel was used to generate a random order for the operating surgeon to perform FH (N=4), SG (N=5), and TR (N=9) experiments. AR (N=6) experiments were performed consecutively as there was not a human operator present to influence performance of the system. Prior to experiments using the AUTO-DALK system, the surgeon was given up to 30 minutes to interact and learn the functionality of the GUI. For FH insertion, an expert surgeon inserted a 20G vertical needle to a target depth under the microscope-integrated B-mode OCT imaging system (Leica Microsystems, Germany), in which the OCT images were hidden from the surgeon. For SG insertion, the surgeon manually inserted the vertical needle with integrated OCT fiber sensor to receive the real-time depth feedback, until the needle tip approached 80%~90% depth of the cornea thickness. For TR insertion, the OCT embedded needle probe was mounted on our AUTO-DALK system, which was then fixed on the rabbit eye by vacuum suction. The surgeon advanced the needle by using the GUI and microscope using the OCT depth sensing feedback. For the AR insertion test, an autonomous control strategy, combined with a corneal layer segmentation algorithm for OCT images, was used to guide the needle probe to the optimal position. A target depth of 100 μm above Descemet's membrane was selected based on our previous work in [Opfermann-2024-design, Wang-2024-live, Wang-2023-common], which aimed for a residual stroma thickness of 10%-20%. This target depth was determined by considering both the corneal thickness and perforation rate from our earlier results. After each insertion test, the surgeon injected air and performed a pneumodissection using a 5 ml syringe. The cornea was dissected and imaged using B-mode OCT to measure the residual stromal tissue. The needle was replaced after two trials. All the microscopic and OCT B-mode images were recorded during the experiments.

Figure 4 and Table S1 in the supplementary materials summarized results from the *ex vivo* study. The average central thickness of these corneas was measured to be 854.7 μm with a standard deviation of 149.2 μm. Perforation of the endothelial layer occurred in a total of 3 cases across all study groups, and blanching of the cornea stroma with successful pneumodissection was observed in other cases. The perforation rates were measured to be 0%, 20.0%, 22.2%, and 0% for FH, SG, TR, and AR, respectively. The incision made by the AUTO-DALK system is smaller and smoother than the others, as detailed in supplemental Figure S3. Compared with FH test, the SG group was statistically more consistent in achieving the ideal needle depth (P = 0.0234), validating the importance of OCT signals in providing real-time depth feedback (Fig. 4a). Perforations are not included in this plot as those samples do not have a measurable distance between the final needle depth and Descemet's membrane. When comparing the SG and TR groups, we see that the robotic system effectively minimized surgical tremors and enabled more accurate needle positioning at target depth (P = 0.0463). It was observed that autonomous needle placement was the most consistent amongst all groups and had an average needle depth with standard deviation of 103.3 ± 4.5 µm. The perforation rate was significantly less for the AR group compared to the SG and TR groups (P < 0.05) and achieved zero perforations despite being nearly 400 µm deeper than the FH approach.

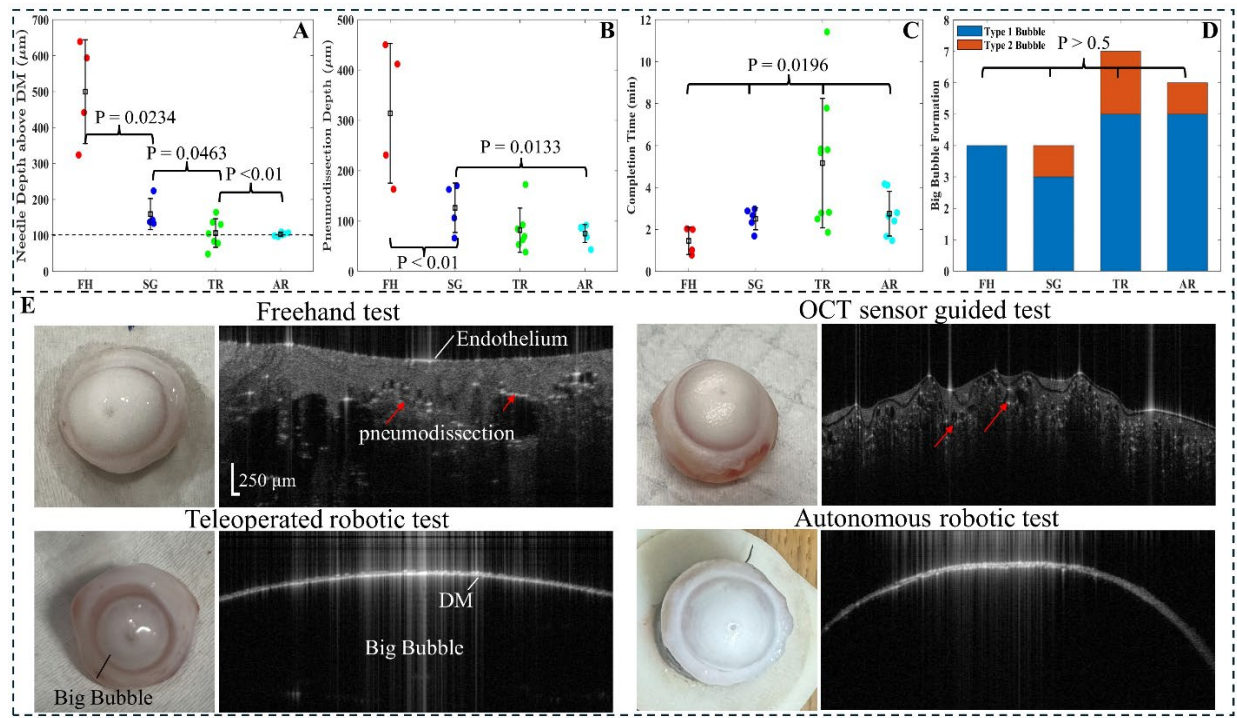

**Figure 4. The results of *ex vivo* big bubble test via FH (n = 4), SG (n = 5), TR (n = 9), and AR (n = 6).** (A) needle depth above Descemet's membrane, (B) pneumodissection depth, (C) task completion time, (D) big bubble formation, (E) representative pneumodissection results of color images with blanching of the corneal stroma and OCT images (endothelial side up) via FH, SG, TR, and AR.

Compared with other approaches, there is a significant variation in the FH pneumodissection depth without OCT depth feedback (P < 0.01) (Fig. 4b). When the surgeon is provided the vertical intraoperative OCT depth sensing needle, the pneumodissection depth improved to 126.4 ± 42.6 µm, up ~22.0% from freehand (SG vs. FH). Like the needle depth, The AUTO-DALK system (AR) achieved a more consistent stromal demarcation depth from pneumodissection of 75.2 ± 16.4 µm compared with SG (P = 0.0133). A notable variation in insertion time was observed across different approaches (P = 0.0196) (Fig. 4c). TR approach required significantly more time than other approaches, primarily due to the need for operator supervision, which involved analyzing OCT M-scans, making slight positional adjustment to the needle probe, and occasional hesitations that was unnecessary in the autonomous mode, adding approximately ~2.5 mins per sample. Interestingly, the completion time for AR was similar between *in vivo* and *ex vivo* testing (P>0.2488), likely due to the significant reduction in corneal thickness of *in vivo* corneas versus *ex vivo* corneas [shimmura-2006-deep]. Although the FH approach yielded the highest the incident of type 1 big bubble formation in *ex vivo* studies with no perforation, approximately 36.7% residual stroma remains, which is beyond the ideal range for effective stromal demarcation in big bubble DALK [Feizi-2016-use]. No statistical difference of big bubble formation was observed among other groups (P > 0.5) (Fig. 4d). Figure 4e illustrates the representative OCT images of stromal demarcation after injecting air with endothelial side up in all study groups. The stroma was fully separated from the Descemet's membrane and a whole bubble was generated in 2 TR tests and 1 AR test. In other perforation-free cases, blanching of the stroma with partial bubble was observed, which were likely caused by endothelium cell loss or shallow needle depth. Furthermore, the NASA task load index was administered following each experiment, and the robotic assisted approach significantly reduced the mental demand, physical demand, and perceived effort compared to the FH and SG guided approaches (See supplemental section *NASA-TLX Analysis* for additional details.)

**TS-U-Net based tissue tracking**

To track the Descemet's membrane layer in real time during needle insertion, we developed an end-to-end segmentation network that incorporates shape and topological constraints, termed the topology-preserving shape-regularized U-Net (TS-U-Net). The entire pipeline consists of a preprocessing step, a customized U-Net architecture, and edge contour extraction, as illustrated in Fig. 5. The basic architecture of the proposed neural network was adapted from U-Net, a model widely used in biomedical image segmentation due to its scalable architecture and multi-resolution analysis [Azad-2024-medical]. The signal to noise ratio (SNR) of the posterior cornea is adversely affected by the thick stroma tissue, especially when the robot stops advancing and OCT needle probe faces a relatively fixed point. We incorporated prior structural characteristics into the training procedure of U-Net to regularize the boundary pixels towards a convex shape [Tajbakhsh-2020-embracing, Mirikharaji-2018-star]. For testing, the positions of epithelium and Descemet's membrane were further extracted from the segmentation map using contour detection and a Kalman filter.

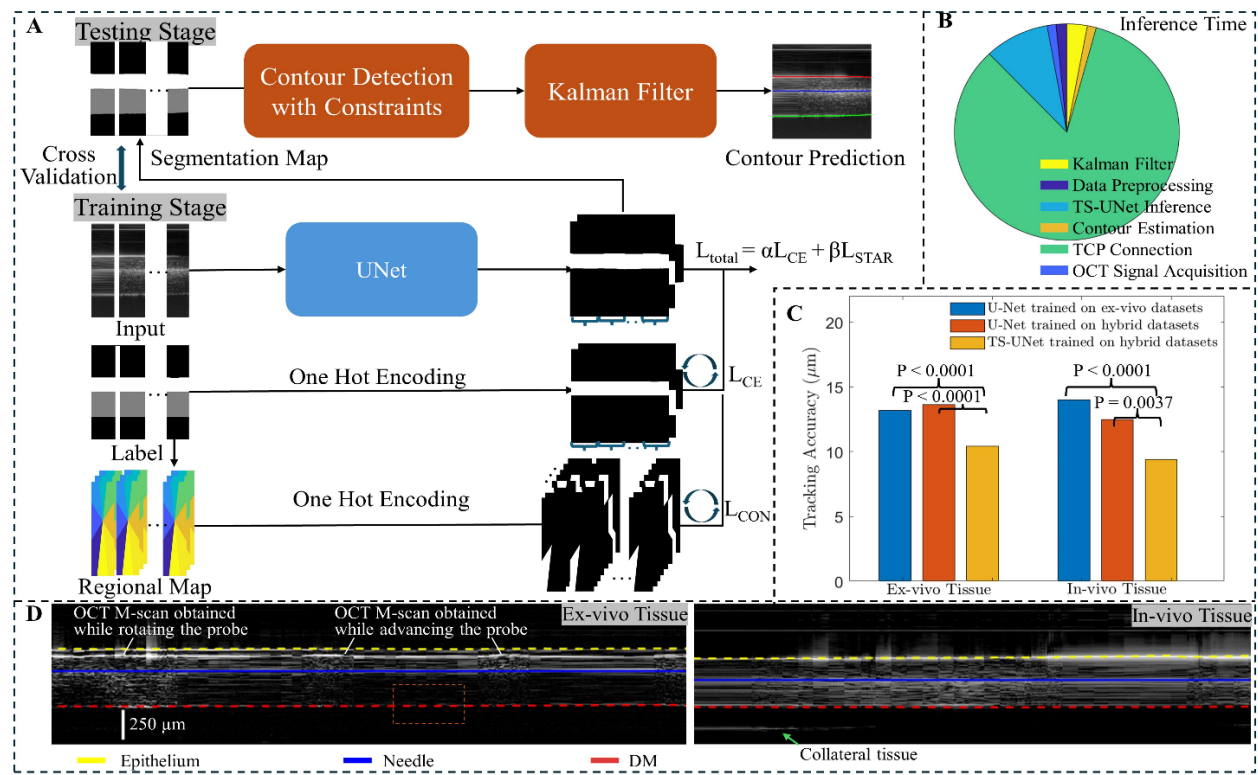

**Figure 5. TS-U-net based tissue tracking.** (A) The proposed pipeline of TS-U-Net based corneal tissue tracker. (B) Inference time of six modules per OCT M-scan (48 OCT A-lines). The total processing time is ~0.15 s. (C) The accuracy test results for the corneal layer tracker on both *ex vivo* and *in vivo* datasets via U-Net trained on *ex vivo* datasets, U-Net trained on hybrid datasets, and TS-U-Net trained on hybrid datasets. (D) The representative segmentation results of the proposed TS-U-Net using *ex vivo* and *in vivo* OCT M-scans.

Fig. 5b summarizes the inference time of six modules per OCT M-scan (48 A-lines), calculated by averaging over 100 loops in Python. The total inference time of TS-U-Net is ~0.15 s, yielding a refresh rate of ~6.64 Hz, which is sufficiently fast to guide the AUTO-DALK robot in practice. The TCP connection step is significantly more time consuming than other modules; however, it can be accelerated by using a smaller image size.

The proposed TS-U-Net was trained and tested by using 2378 OCT M-scans (1024 × 1024 pixels) from 62 *ex vivo* corneal tissues and 793 OCT M-scans (1024 × 1024 pixels) from 13 *in vivo* corneal tissues acquired during big bubble DALK procedures. Ground truth annotations were reviewed for accuracy by an ophthalmologist. Five-fold cross-validation was employed across the whole dataset to mitigate overfitting issue and reduce bias. No OCT M-scan from the same corneal tissue was utilized in both training and testing folds. An ablation study was conducted on the TS-U-Net and two basic U-Nets: one trained on the *ex vivo* dataset, and the other trained on both *ex vivo* and *in vivo* datasets. All the neural network hyperparameters were optimized prior to evaluation. It should be noted that the needle depth was manually typed into the GUI during actual use, as the relative distance between the needle and fiber tip remained unchanged in all insertion tests. The results showed in Fig. 5c demonstrate that our proposed TS-U-Net is more robust to various noises in OCT M-mode images. On the *ex vivo* dataset, our model improves the tracking accuracy to $10.41 \pm 7.48$ μm (P<0.001). On the *in vivo* dataset, accuracy improves to $9.39 \pm 5.82$ μm (P<0.001). Incorporating *in vivo* images into the training dataset significantly enhances the robustness of the models for *in vivo* study. Fig. 5d indicates that our model can accurately segment tissue layers from OCT M-scans even in cases of low SNR or the presence of ghost images.

## DISCUSSION

In this paper, we demonstrated an autonomous surgical robotic system capable of precise needle insertion and deep stromal pneumodissection for big bubble DALK procedure. To the best of our knowledge, it has performed the first vertical needle placement and the first autonomous robotic big bubble procedure on *in vivo* live animal eye models. Our results demonstrated that the AUTO-DALK system outperformed both expert surgeon and other study groups across key metrics, including the needle depth above Descemet's membrane, pneumodissection depth, and task completion time. We believe this AI-driven robotic approach can improve the consistency and success rate of DALK, while reducing the risk of graft rejection.

A fiber-optic distal sensor integrated downward viewing needle probe maximizes the backscattered OCT signals and allows better Descemet's membrane tracking throughout a long sensing range (~3 mm). The sensor not only acquires one-dimensional OCT A-mode and M-mode signals but can also collect circumferential cross-sectional images of internal corneal tissues by spinning the needle probe through the rotary module. Compared with 2D imaging solutions, the sensor does not require a complex needle path planning along two dimensions, which is easy to operate especially with our user-friendly interface. In the clinical environment, the excessive ocular movements and random deformation of the tissue could be a surgical challenge, which could increase the risk of perforation. The *in vivo* study validated that the lightweight and compact AUTO-DALK system generated a powerful suction to stabilize itself on the eye, and therefore it was immune to the physiological motions of animals. The vertical needle that integrated the OCT fiber provided a continuous and precise depth feedback of corneal tissue layers, which enabled a successful and consistent surgical outcome in live models.

Previous studies of horizontal approaches have shown that a successful pneumodissection strongly depends on the final insertion depth, normally 80%~90% of corneal thickness. However, this target range needs to be fine-tuned for the vertical approach. When using our novel vertical approach, the *in vivo* experiments showed that the stromal pneumodissection depth was ~46.7 μm greater than the needle depth. Additionally, if the insertion depth is greater than ~44.6 μm (87.9%) from the Descemet's membrane, there is an increased risk of perforation during pneumodissection. We attribute this finding to the large downward injection pressure used in our vertical system, as opposed to a horizontal pressure that would be present in the clinical standard of care. In the experiment, we encouraged the operator to explore a more appropriate target range for needle depth by utilizing the real-time distal sensing on both *in vivo* and *ex vivo* models. We believe these results will help our surgical robot to perform more accurate and consistent pneumodissection in the future. We chose to use an absolute distance in μm for the insertion depth rather than a percentage of corneal thickness because the variation in corneal thickness among our live rabbit models is minimal, with a standard deviation of 24.2 μm. Additionally, specifying the target depth in microns within the GUI simplifies comparisons for subsequent evaluations without affecting test accuracy.

In the *ex vivo* study (N = 28), the AUTO-DALK system formed 3 big bubbles and achieved sparing of Dua's layer. In the remaining 17 robotic and 12 manual trials, the corneal stroma was filled with small bubbles after pneumodissection as shown in Fig. 3 and 4. Research has shown that either scenario will not affect the Descemet's membrane's separation or surgical outcome [Sugita-1997-deep]. However, an additional lamellar dissection is then required in the latter situation, with a clinical incidence of 10%-50% [Cheng-2013-comparison]. In the 3 cases where a clear big bubble separation occurred, we measured the thickness of the residual posterior cornea and confirmed that Dua's layer was preserved, with Type 1 big bubble formation occurring after pneumodissection. This was validated by comparing postoperative OCT images to a structural model of the cornea [Saikia-2018-basement, Riau-2012-reproducibility]. Type 1 big bubbles are desirable for the DALK procedure, as Dua's layer provides additional strength to the transplanted eye [Zaki-2015-deep]. The *ex vivo* results suggest there is potential for our AUTO-DALK system to form a desired Type 1 bubble with cleavage separation between Dua's layer and posterior stroma, which can further reduce the operating time and improve the repeatability of this big bubble technique.

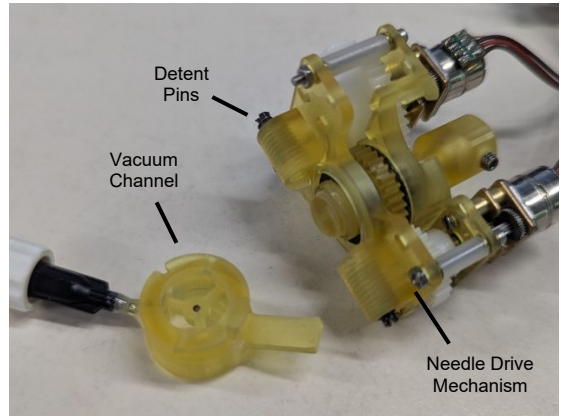

Additionally, the AUTO-DALK system demonstrated significantly more ideal and consistent needle placement, achieving a similar incidence of type 1 big bubble formation to manual insertion. To improve results between AR1 and AR2 study groups in our *in vivo* study, we address two key concepts. First, a two-piece needle alignment system was integrated into the AUTO-DALK robot to minimize needle runout during needle insertion as shown in Fig. 6. The two-piece design allows the surgeon to center the base of the AUTO-DALK at the apex of the cornea and attach the base independently with the vacuum syringe. After the base is attached, the top half of the AUTO-DALK is mounted onto the base, with the needle inserted into the guide bearing sleeve. The two halves are held in position with a detent pin.

**Figure 6. Two-piece AUTO-DALK robot.** Illustration of the two-piece AUTO-DALK design. The vacuum channel can be centered and selectively attached to the cornea, followed by attachment of the needle drive mechanism to the vacuum base.

Second, we improve our AI-based segmentation and tissue tracking model by implementing geometric constraints, and including additional *in vivo* training data. This enabled fine adjustment of the tissue tracking algorithm to more accurately identify the Descement's Membrane in *in vivo* tissue. Therefore, the postmortem thickening of the stromal layer or swelling of the eye does not compromise the robustness of our proposed robotic system.

Despite these achievements, some limitations remain in this study. First, approximately 30% of the OCT embedded needles were compromised during needle probe manufacturing due to insufficient polishing, especially when immersed in vitreous media (e.g., tears, mucus), which can lead to OCT signal degradation. It is essential to test the signal-to-noise ratio (SNR) of OCT signals and replace the needle probe, if necessary, before each test. In future work, we aim to optimize and standardize the manufacturing process to reduce the rejection rate. Although much of the workflow is automated, surgeons still need to manually inject air to create pneumodissection. Our group previously proposed a robotic injection system, and we plan to integrate it into our next-generation AUTO-DALK robot to further minimize human intervention, moving us closer to a fully autonomous solution [Kaluna-2024-A]. Additionally, while rabbits are a common model for *in vivo* corneal experiments, there are biomechanical differences between rabbit and human tissues [Ojeda-2001-three]. The system should undergo further evaluation using cadaver human corneas before progressing to survival studies [draelos-2019-optical].

## MATERIALS AND METHODS

**Robotic Platform.** We have designed the vertical needle insertion robot illustrated in Fig. 7. The robot consists of two symmetric micro stepper motors that independently rotate a central needle housing. When the right motor rotates, it turns a gear with an internal screw. If the left motor is not energized, the needle housing is prevented from rotating, replicating the motion of a lead screw such that the needle has pure translation in the vertical axis. If the left motor is energized at the same relative speed as the right motor, the needle housing and lead screw gear turn at the same effective speed which enables pure rotation of the needle. Finally, if the left gear rotates either faster or slower than the relative speed of the right motor, a screw motion is induced on the needle housing which enables the needle to simultaneously rotate and translate with up to 5N of thrust. The vertical robot was designed using SolidWorks computer aided design software and prototyped using a microArch 3D printer with 40 μm resolution (Boston Micro Fabrication, Maynard, MA). Open loop average positional deviation of the system was measured to be 3.88 μm following ISO standards (See supplemental materials for additional details). Encoded motors were not deemed necessary for this robot design, as OCT depth sensing feedback and image guidance is used to perform closed loop control of the needle tip. The robot also integrates a vacuum chamber and commercial spring-loaded syringe to attach the robot to the eye with vacuum pressure. The same geometry used for commercial vacuum trephines is used for this design such that the administered vacuum pressure does not exceed clinically accepted safety limits. The vertical needle can be inserted into the needle housing and fixated using a set screw for quick change between experimental trials. A 90-degree mirror is mounted to direct light through the side of the robot such that a downward facing surgical microscope can be used to visualize the needle as it is inserted into the cornea. Control of the stepper motors is performed using an Arduino UNO microcontroller with two Easy Driver (Sparkfun, Boulder, CO) motor drivers. The Arduino is interfaced with the GUI using a ROS package.

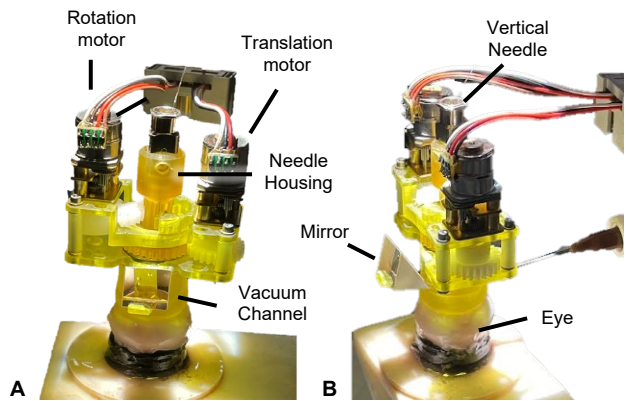

**Figure 7. Eye mountable AUTO-DALK robot.** (A) Front view of the AUTO-DALK robot vacuum mounted to a rabbit whole eye. (B) side view of the AUTO-DALK robot with attached prism mirror to enable direct visualization with surgical microscope.

**Imaging System.** The in-house built OCT system is shown in Fig. 2a [Guo-2020-demonstration]. It is comprised of a swept-source OCT engine (Excelitas Technologies, Billerica, USA), an optical circulator, a Camera-Link frame grabber, a single-mode fiber (1060XP, Coating Diameter: 245 μm, Cladding Diameter 125 μm, NA: 0.14), and a desktop. The system has a center wavelength of 1,060 nm, with a sweeping bandwidth of 100 nm, sweeping rate at 100 kHz, and an output power of 2 mW. The axial resolution of the system is ~6 μm in air and ~4.5 μm inside the tissue. The maximum sensing depth is ~3.7 mm in air and ~2.8 mm inside the tissue. The reference beam and sample beam share the same optical path, so there is no need for removal of polarization noise or dispersion compensation.

As shown in Fig. 8a, the OCT fiber probe is mounted 500 μm from the tip of the razor needle and has an acceptance cone for reflected OCT signals. The fiber is epoxied to a customized 20-gauge needle (Vita Needle Company, Needham, USA, ID: 0.0235″, OD: 0.0355″, length: 1.20″) where the needle tip is ground to a razor edge so that rotation of the needle can cut tissue as it is inserted into the stroma (Fig. 8b). To protect the fiber tip, the OCT fiber sensor is encapsulated within another stainless-steel tubing (ID: 0.013″, OD: 0.016″) by epoxy adhesives (Fig. 8c).

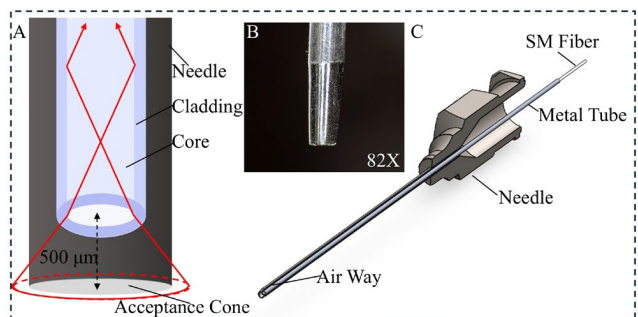

**Figure 8. The design of OCT needle probe.** (A) Illustration of the acceptance angle of a fiber, (B) 82X microscope image of the 20G razor edge needle, (C) cross-section view of the OCT needle probe.

The extra epoxy that protrudes out of the tube is polished to an 8-degree angle. This tubing can provide support and stability for the fragile fiber, especially when the needle starts to advance or rotate. it can also prevent signal attenuation of the reference beam due to the watery fluid or stroma tissue that remain inside the needle in the needle insertion test. The 20G needle size leaves enough space for pneumodissection. Compared with commonly used beveled needles, blunt needles create higher peak forces and cause larger soft tissue deformation when they start to pierce a tissue [Van-2012-needle], which is up to 2.5 mm depth on the corneal tissue. Encasing the fiber probe inside the needle can measure the needle tip's depth more accurately. As shown in Fig. 8a, the backscattered light from the needle tip can propagate in fibers with a suitable offset and fiber acceptance angle of 8°. Furthermore, due to a relatively shorter travel length of needle, this vertical sensor placement has higher SNR and lower attenuation of Descemet's membrane on OCT A-line signals than conventional lateral approaches with an angle of ~15°.

The OCT spectrum data is captured by the frame grabber and sent to the desktop for further processing, which includes filtering, zero-padding, and fast Fourier transform (FFT). To reduce speckle noise, 256 A-line signals are averaged over an integration time of 2.56 ms. The background signals are subtracted from the raw OCT signals. Moreover, the measured optical length needs to be calibrated against the sample's refractive index [Huang-1991-optical], given as,

$$l_g = \frac{l_o}{n_s}, \qquad (1)$$

where $l_g$, $l_o$, and $n_s$ denote the geometric length after calibration, the measured optical length, and the refractive index of tissues. Once the needle touches the epithelium, the refraction correction will be applied to OCT M-scans to increase depth sensing accuracy and avoid perforation. All above steps are written in C++ and C# and implemented on GPU (NVIDIA RTX 4070 Ti Super). Instead of displaying A-line signals on the user interface, OCT M-scans are shown to surgeons, which can provide a time series of A-lines with more anatomical information. Fig 2b shows the OCT M-scan image and corresponding A-line signal.

**TS-U-Net based Tissue Tracking.** Fig. 5 demonstrates the framework of our proposed tissue tracking model. For data preprocessing, 48 raw OCT A-line signals were received through TCP at a time, which were treated as a single OCT M-scan. To reduce data variability, the input M-scan was standardized, and the pixel values were linearly scaled between -3 and 4. Each M-scan was then normalized to the [0,1] range. Symmetric padding with a size of 8 was applied to preserve image details and maintain the continuity of patterns, which was crucial given the small receptive field of our TS-U-Net ($7 \times 3$). U-Net's fully convolutional structure and skip connection enable accurate localization of object boundaries, even with small training datasets [Wei-2022-deep]. The basic architecture of TS-U-Net was developed from [Ronneberger-2015-u], allowing it to identify relevant features and predict the segmented maps of corneal tissues. Our proposed model consists of a contracting path and an expansive path. For each individual stream, a $7 \times 3 \times N$ matrix was extracted, where $7 \times 3$ represents the spatial neighborhood size and N is the sample size. The Kaiming weight initialization was applied to the activation function of ReLU [He-2015-delving]. Same padding was implemented for all pooling layers to prevent the resolution reduction, and a dropout unit with a rate of 0.5 was used at the bottleneck to avoid overfitting. Additional constraints including the shape regularization, were applied on predicted segmentation masks to produce more realistic outputs, particularly in scenarios with lower SNR.

During training, region maps were automatically generated from ground truth maps, which included the center $c$ of each object along with 8 ROIs surrounding $c$. In addition to the cross-entropy loss $L_{ce}$, an extra convex shape loss $L_{con}$ was employed, constraining any point between the center $c$ and an interior point to remain within the interior. It helps eliminate small holes and produce smoother segmentation maps, especially when the structural features of the posterior corneal signals are limited. Scaling factor $\alpha = 1$ and $\beta = 0.12$ were applied to $L_{ce}$ and $L_{con}$ to enhance the model's generalization ability. Notably, we did not need to compute the regional map or object centers for testing, as the prediction map was generated directly through the forward direction of the TS-U-Net. The segmentation map was convolved with vertical derivative filters to detect the positions of the epithelium and Descemet's membrane. Possible edge points were identified if the gradient magnitude at a pixel exceeds a specified threshold. The final edge line of the epithelium was determined by computing the standard deviation of each potential line. In general, the segmentation results for the epithelium are better than those for the Descemet's membrane, as the epithelium is closer to the fiber probe and benefit from stronger OCT signals. Therefore, we imposed an area constraint on the Descemet's membrane location, specifying that the distance from the epithelium to the Descemet's membrane should range from [0,400] μm for *in vivo* tissues, and [0,1100] μm for *ex vivo* tissues.

To further enhance the tracking accuracy of the corneal layers, we employed the Kalman filter that configured with a state transition matrix of $F = 1$, an observation matrix of $H = 1$, a process noise covariance of $Q = 1 \times 10^{-5}$ and observation noise covariance of $R = 1$. This configuration helps to get a good tradeoff between the smoothing effect and responsiveness of Kalman filter. We implemented a sliding window technique to further refine the observation input. Initially, for the first 50 data points, the standard Kalman update is applied to establish a reliable state estimate. As additional data accumulates, the approach adapts by using a weighted average of the most recent 50 points with the weight of 0.7 and the preceding 50 points with the weight of 0.3 as the observation input. Under this circumstance, it remains responsive to new data while maintains the stability from prior observations, providing a smooth and consistent edge line of the corneal layers. Specifically, the state estimate $\hat{x}_{k|k}$ is updated recursively using the Kalman gain $K_k$. The prediction step is given by $\hat{x}_{k|k-1} = F\hat{x}_{k-1|k-1}$, and the update step incorporates the weighted observation: $\hat{x}_{k|k} = \hat{x}_{k|k-1} + K_k(z_k - H\hat{x}_{k|k-1})$.

**Autonomous Control Strategy.** The overall control strategy is illustrated in Fig. 2, and it is based on a feedback control loop. The real-time OCT interferogram from the tool tip is captured and sent to a frame grabber, and then to a desktop for tracking the position of the corneal layers. The TS-U-Net updates the relative distance between Descemet's membrane and needle tip using its previous and current predictions. The system also records all robot movements and transmits this data to the GUI for subsequent execution. The control loop operates under two conditions: the needle depth must be shallower than the target depth, and the total travel distance must not exceed the maximum travel length of motors. If statement is satisfied, the AUTO-DALK system generates two motion primitives for advancement and one for rotation, sending one command in each cycle. Commands are sent to an Arduino UNO which utilizes two stepper motor drivers to control motion of the needle. If the distance between needle depth and target depth is less than 100 μm, it will advance the needle at half of the preset step size to avoid perforation.

**Graphical User Interface.** The Graphical User Interface (GUI) was developed in C++ and should be used in Ubuntu 22. It utilizes ROS2 to pass data and communicate with the robot, and Qt to render the application and facilitate its functions. The GUI supports manual and autonomous insertions of the needle by providing buttons to move the needle up/down by some user-defined step size (1-150 μm). The needle can be set to continuously rotate as well. Status lights and an error readout are included here in case something goes wrong. Next, the OCT M-Mode display has a pause button, adjustable time range (1-9 s), and adjustable distance range (256-3000 μm). The distance range can also be automatically set such that Descemet's membrane is always within view. The locations of Descemet's membrane and epithelium are determined by the tissue tracking algorithm and sent over TCP to the GUI computer. A ROS2 node picks up and publishes the TCP message, and the GUI (running in a different node) subscribes to those messages. Descemet's membrane and the epithelium are displayed on the OCT M-mode as toggleable traces, and the user can enable a target-depth trace at some percent of the distance between the two. Signal preprocessing is employed to enhance data visualization in GUI. To improve the image contrast, histogram equalization was applied to OCT M-scan, which has a size of 1024 × 48.

**Statistical analysis.** Statistical analysis was conducted using MATLAB. The Kruskal-Wallis test was performed for the tissue tracking accuracy. To evaluate the consistency of surgical outcomes, Levene's test for variance was applied to assess the variance of the surgical outcomes, including the needle depth, pneumodissection depth, and task completion time. Additionally, Levene's test followed by a t-test was conducted to assess the mean values of the performance metrics. The analysis of variance test (ANOVA) was used to evaluate the task completion time and big bubble formation. P values are reported for all tests, with a significance level set at $P < 0.05$.

[11] phantoms, in 2015 IEEE International Conference onRobotics and Automation (ICRA), (IEEE, 2015), pp. 1202–1209.

[12] Lu-2021-super: Lu, Jingpei, et al. "Super deep: A surgical perception framework for robotic tissue manipulation using deep learning for feature extraction." 2021 IEEE International Conference on Robotics and Automation (ICRA). IEEE, 2021.

[13] Saeidi-2022-autonomous: Hamed Saeidi, Justin D Opfermann, Michael Kam, Shuwen Wei, Simon Léonard, Michael H Hsieh, Jin U Kang, and Axel Krieger. Autonomous robotic laparoscopic surgery for intestinal anastomosis. Science robotics, 7(62):eabj2908, 2022.

[14] MacDonald-2005-learning: MacDonald JD. Learning to perform microvascular anastomosis. Skull Base. 2005;15(3):229–240. doi: 10.1055/s-2005-872598.

[15] Nasseri-2013-surgery: M. A. Nasseri et al., "The introduction of a new robot for assistance in ophthalmic surgery," Annu. Int. Conf. IEEE Eng. Med. Biol. Soc. IEEE Eng. Med. Biol. Soc. Annu. Int. Conf., vol. 2013, pp. 5682–5685, 2013, doi: 10.1109/EMBC.2013.6610840.

[16] Zhao-2023-cannula: Zhao Y, Jablonka AM, Maierhofer NA, Roodaki H, Eslami A, Maier, M, Nasseri MA, Zapp D. Comparison of Robot-Assisted and Manual Cannula Insertion in Simulated Big-Bubble Deep Anterior Lamellar Keratoplasty. Micromachines (Basel). 2023 Jun 16;14(6):1261. doi: 10.3390/mi14061261. PMID: 37374846; PMCID: PMC10301424.

[17] Cheon-2017-motorized: Cheon GW, Gonenc B, Taylor RH, Gehlbach PL, Kang JU. Motorized microforceps with active motion guidance based on common-path SSOCT for epiretinal membranectomy. IEEE/ASME Trans. Mechatronics 22, 2440–2448 (2017).

[18] Lee-2021-cnn: Lee S, Kang JU. CNN-based CP-OCT sensor integrated with a subretinal injector for retinal boundary tracking and injection guidance. J. Biomed. Opt. 26, 068001 (2021).

[19] Huang-2012-motion: Huang Y, Liu X, Song C, Kang JU. Motion-compensated hand-held common-path Fourier-domain optical coherence tomography probe for image-guided intervention. Biomed. Opt. Express 3, 3105–3118 (2012).

[20] Taylor-1999-steady: R. Taylor et al., "A Steady-Hand Robotic System for Microsurgical Augmentation," in Medical Image Computing and Computer-Assisted Intervention – MICCAI'99.

[21] Mitchell-2007-steady: B. Mitchell et al., "Development and Application of a New Steady-Hand Manipulator for Retinal Surgery," in Proceedings 2007 IEEE International Conference on Robotics and Automation, Apr. 2007, pp. 623–629. doi: 10.1109/ROBOT.2007.363056.

[22] Blum-2009-lasik: Blum, Marcus, Kathleen Kunert, Annika Gille, and Walter Sekundo. "LASIK for myopia using the Zeiss VisuMax femtosecond laser and MEL 80 excimer laser." Journal of Refractive Surgery 25, no. 4 (2009): 350-356.

[23] Sugita-1997-dalk: J. Sugita and J. Kondo, "Deep lamellar keratoplasty with complete removal of pathological stroma for vision improvement.," Br. J. Ophthalmol., vol. 81, no. 3, pp. 184–188, Mar. 1997.

[24] Cheng-2013-comparison: J. Cheng, X. Qi, J. Zhao, H. Zhai, and L. Xie, "Comparison of penetrating keratoplasty and deep lamellar keratoplasty for macular corneal

2023.Ojeda-2001-three: Ojeda JL, Ventosa JA, Piedra S. The three-dimensional microanatomy of the rabbit and human cornea: A chemical and mechanical microdissection-SEM approach. J. Anat. 199, 567–576 (2001).

[51] Draelos-2019-optical: Draelos M, Tang G, Keller B, Kuo A, Hauser K, Izatt JA. Optical coherence tomography guided robotic needle insertion for deep anterior lamellar keratoplasty. IEEE Trans. Biomed. Eng. 67, 2073–2083 (2019).

[52] Guo-2020-demonstration: Guo S, Sarfaraz NR, Gensheimer WG, Krieger A, Kang JU. Demonstration of optical coherence tomography guided big bubble technique for deep anterior lamellar keratoplasty (DALK). Sensors 20, 428 (2020).

[53] van-2012-needle: van Gerwen DJ, Dankelman J, van den Dobbelsteen JJ. Needle–tissue interaction forces: A survey of experimental data. Med. Eng. Phys. 34, 665–680 (2012).

[54] Huang-1991-optical: Huang D, Swanson EA, Lin CP, Schuman JS, Stinson WG, Chang W, Hee MR, Flotte T, Gregory K, Puliafito CA, et al. Optical coherence tomography. Science 254, 1178–1181 (1991).

[55] Wei-2022-deep: Wei S, Kam M, Wang Y, Opfermann JD, Saeidi H, Hsieh MH, Krieger A, Kang JU. Deep point cloud landmark localization for fringe projection profilometry. J. Opt. Soc. Am. A 39, 655–661 (2022).

[56] Ronneberger-2015-u: Ronneberger O, Fischer P, Brox T. U-net: Convolutional networks for biomedical image segmentation. Med. Image Comput. Comput. Assist. Interv. (MICCAI) 234–241 (2015).

**Acknowledgments:** The authors would like to thank Dr. Ian Dobbie and the Integrated Imaging Center at Johns Hopkins University for their support in micro 3D printing the AUTO-DALK robot.

**Funding:** Research reported in this paper was supported by the National Institute of Biomedical Imaging and Bioengineering of the National Institutes of Health under award number 1R01EY032127.

**Author contributions:**
Conceptualization: YW, JO, WG, AK, JUK
Methodology: YW, JO, JY, HY, JK, WG, AK, JUK
Software: YW, JY, HY, RB
Investigation: YW, JO, JY, HY, JK, WG, AK, JUK
Visualization: YW, JO, JK, RB
Data Curation: YW, JO, JY, HY, JUK
Formal Analysis: YW, JO, JK, RZ
Funding acquisition: WG, AK, JUK
Supervision: WG, AK, JUK
Writing – original draft: YW, JO, JY, JK, AK, JUK
Writing – review & editing: YW, JO, WG, AK, JUK
**Competing interests:** Authors declare that they have no competing interests.
**Data and materials availability:** All data are available in the main text or the supplementary materials.

## Supplementary Materials

Materials and Methods
Fig. S1 Figure S1. *In vivo* experimental setup.
Fig. S2. *Ex vivo* experimental test setups.
*Fig.* S3. *Ex vivo* OCT images of rabbit eyes.
Fig. S4. Calibration step for the proposed AUTO-DALK system.
Fig. S5. Tabulated results of the NASA task load index questionnaire.
Table S1. Ex vivo and in vivo pneumodissection results.

# Supplementary Materials and Methods

## AUTO-DALK positional accuracy and repeatability

To measure the open loop accuracy and repeatability of the system, we first performed a calibration step to measure the relationship between the commanded distance and measured travel distance of the AUTO-DALK system. The calibration was performed by attaching a vertical needle with embedded OCT probe to the AUTO-DALK system, and positioning the needle perpendicular to a target metal plate 2 mm away. The motor was commanded to move 50 μm towards the metal target for a total of 20 times, and the travel distance was measured by the OCT A-line signal. This test was repeated for a total of 6 times, and the average travel distance for each step was plotted in a scatter plot as shown in supplemental Figure S4. The relationship between the travelled and commanded distance was linear following the equation (1):

$$y = 0.96x \quad \quad (SEQ1)$$

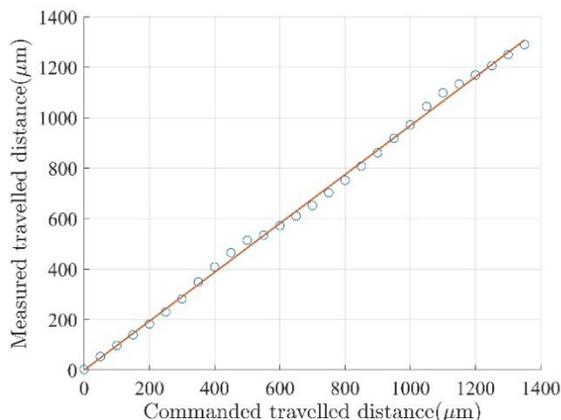

**Figure S4. Calibration step for the proposed AUTO-DALK system.** The relationship between the measured travelled distance y and commanded travelled distance x was linear following the equation y = 0.96x.

Where y is the measured travel distance, and x is the commanded travel distance.

The AUTO-DALK robot was then evaluated for open loop positing accuracy following ISO 230-2:2014. To complete the test, the system was directed to 5 or more target positions $p_i$, given as follows,

$$p_i = (i-1)p + r, \quad \quad (SEQ2)$$

where $i$ is the target position index, $p$ is the nominal interval between each point, and $r$ is a random value within 30% of $p$. Ten positions were generated by Microsoft Excel that were distributed randomly across 1.5 mm travel length and were converted using supplemental Equation (1) to minimize error between commanded and actual positions. The motor was then commanded to reach these target positions from both directions using the above experiment setup. This test was repeated for 5 times. Based on ISO 230-2:2014, the positional deviation is defined as the difference between measured position and target position in either direction; the positional repeatability is defined as four times the maximum standard deviation (σ) of positional deviations; the positional accuracy is defined as the average positional deviations plus or minus 2σ. The computed average positional deviation, positional repeatability, and positional accuracy of the system was 3.88 μm, 53.30 μm, and 57.21 μm, respectively.

## OCT Accuracy Calculation

We manually annotated the position of the epithelium, Descemet's membrane, and needle using MATLAB and ImageJ to create the ground truth mask, which was then converted into one-hot-encoded matrix for training. To test our deep learning model, we extracted the ground truth coordinates of the target layers from these categorical masks using our proposed contour detection algorithm. Meanwhile, TS-U-Net predicted the coordinates of corresponding objects through a combination of the neural network, contour detection algorithm, and Kalman Filter. Tracking accuracy is defined as the absolute distance in microns between the predicted coordinates of target layers and ground truth.

## *In vivo* Animal Protocol

All animal testing was completed with Institutional Animal Care and Use Committee Approval (Protocol 00002276(a)) at Dartmouth-Hitchcock Medical Center in Lebanon, NH. A total of ten rabbits across two surgical dates underwent pneumodissection of the deep stroma in a simulated deep anterior lamellar keratoplasty procedure. Four of the rabbits had pneumodissection performed with the AUTO-DALK robot under teleoperation (n = 8 eyes), and six rabbits had pneumodissection performed with the AUTO-DALK robot under autonomous control (n = 11 eyes). The autonomous experiments were performed in two groups AR1 (n = 5 eyes) and AR2 (n = 6 eyes). Each rabbit was fully anesthetized prior to any surgical procedure by intramuscular injection in the epaxial muscles or rear leg. Pain was relieved a combination of topical anesthetic drops on the ocular surface and subcutaneous injection. Animals were intubated and anesthesia was maintained with vaporized isoflurane first administered a ventilator. Pentobarbital was administered through an ear vein. Following the induction of anesthesia, the eye was prepped and draped. A lid speculum was placed in the eye to hold open the eyelid, and the AUTO-DALK was attached to the apex of the cornea using a vacuum syringe. Following the procedure, the rabbits were euthanized with an overdose of pentobarbital. Death was confirmed by checking for sustained failure of spontaneous respirations efforts, heart tones, lack of motor response to pain stimuli, and fixed pupils. Bilateral thoracotomy was also performed as a secondary procedure. A limited necropsy was performed to remove the cornea for evaluation under OCT imaging.

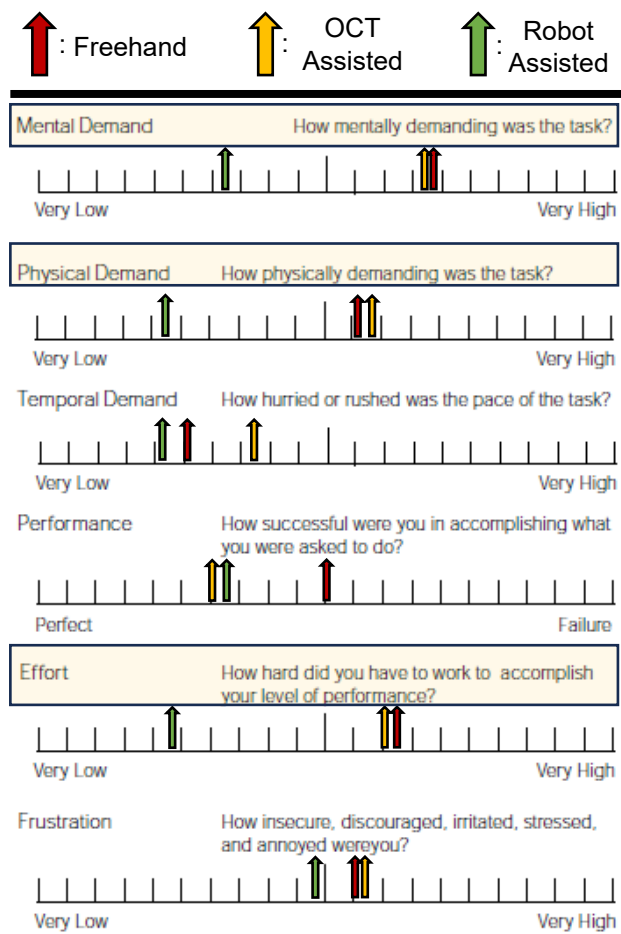

## NASA-TLX Analysis

Following each *ex vivo* trial involving a human operator {freehand (FH), sensor guided (SG), and teleoperated (TR)}, the operating surgeon was administered a NASA-TLX questionnaire to measure the respective task load index for the respective trial. Using a scale from 1-20, the questionnaire quantifies the relative Mental Demand (How demanding was the task), Physical Demand (How physically demanding was the task), Temporal Demand (How hurried or rushed was the task), Performance (How successful were you in accomplishing what you were asked to do), Effort (How hard did you have to work to accomplish your level of performance), Frustration (How insecure, discouraged, irritated, stressed, and annoyed were you). Responses were averaged across trials for each study group respectively, and then plotted on the TLX questionnaire as illustrated in supplemental Figure S5. A student's t-test was used to measure statistical differences between responses for each study group. From the analysis, it was found that the robotic assisted approach (TR) significantly reduced mental demand (7.5 vs. 12.5 and 12.0, P < 0.05), physical demand (5.5 vs. 12.25 and 12.75, P < 0.05) and perceived effort (5.75 vs. 12.5 and 12.0, P < 0.05) compared to the FH and SG groups respectively indicating strong preference of the surgeon to use a robot guided needle insertion approach for the DALK procedure. All other task load index metrics were not found to be statistically significant in this study.

**Figure S5. Tabulated results of the NASA task load index questionnaire.** The NASA TLX was administered during *ex vivo* experiments. Average values for freehand, sensor guided, and teleoperated are shown by red, yellow, and green arrows

**Supplemental Figures**

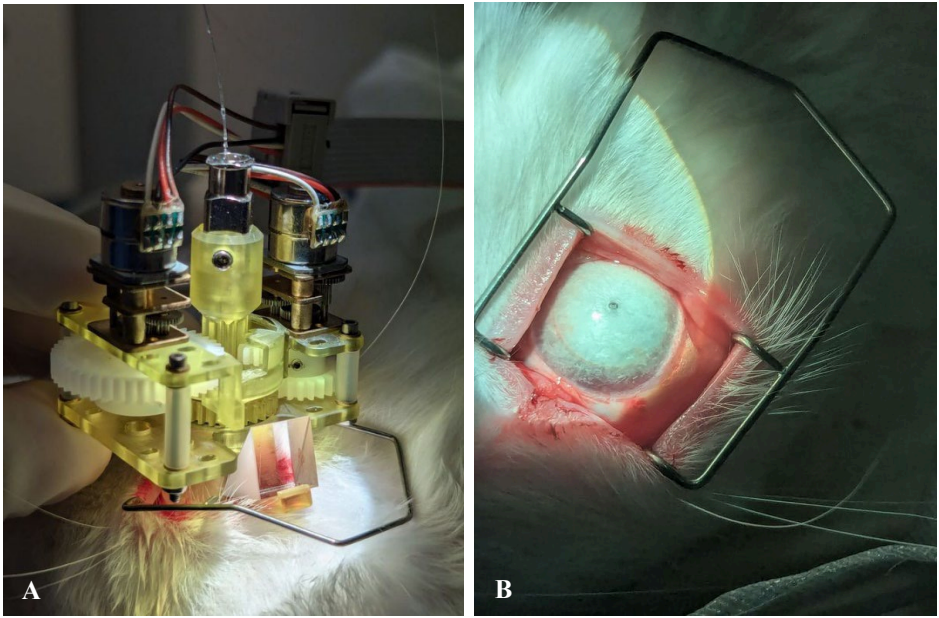

**Figure S1.** *In vivo* **experimental setup.** (A) The AUTO-DALK system mounted to the cornea of a rabbit model. (B) The resulting pneumodissection after needle insertion from the AUTO-DALk system.

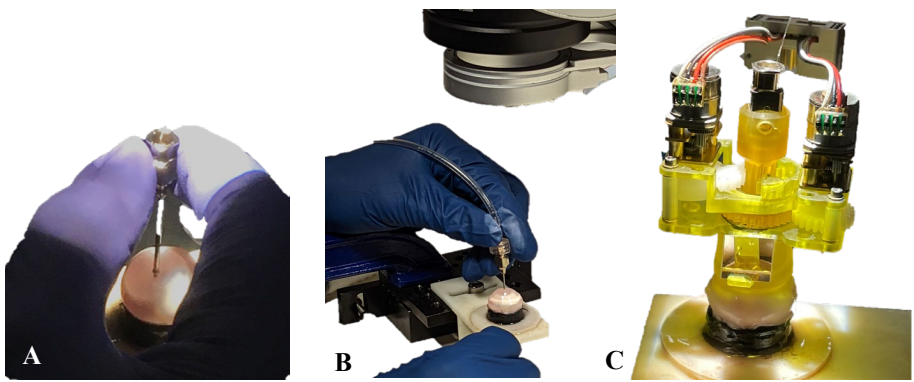

**Figure S2.** *Ex vivo* **experimental test setups.** (A) Freehand surgical approach, (B) OCT sensor guided approach, (C) Robotic setup for teleoperated and autonomous experiments.

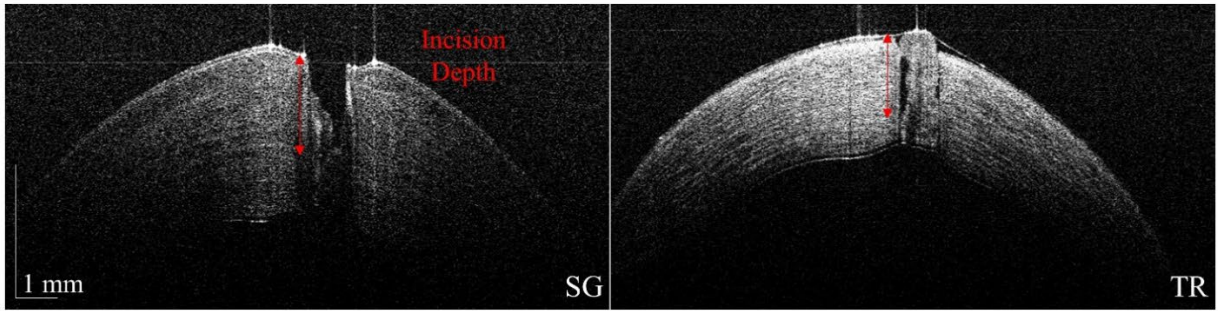

**Figure S3. *Ex vivo* OCT images of rabbit eyes.** Postoperative OCT images showing the needle trajectory before pneumodissection for manual (SG) and robotic (TR) insertion tests.

## Supplemental Tables

**Table S1. Ex vivo and in vivo pneumodissection results.** The performance metrics for evaluating the surgical outcomes for *in vivo* and *ex vivo* experiments, including the perforation rate, needle depth, pneumodissection depth, the incident of type 1 big bubble formation, and task completion time.

|  |  | App. | N | Perforation n | Perforation % | Needle Depth | Pneumodissection Depth | Incident of Type 1 BB | Completion Time |
|---|---|---|---|---|---|---|---|---|---|
| *In vivo* | Robot Driven | TR | 8 | 2 | 25.0% | 97.0 ± 20.3 µm | 57.7 ± 21.6 µm | 83.3% | 227.8 ± 104.1 s |
| | | AR1 | 5 | 1 | 20% | 108.6 ± 4.7 µm | 55.8 ± 18.4 µm | 75.0% | 182.7 ± 29.4 s |
| | | AR2 | 6 | 1 | 16.7% | 102.4 ± 4.8 µm | 51.6 ± 12.5 µm | 60.0% | 128.2 ± 40.4 s |
| *Ex vivo* | Manual | FH | 4 | 0 | 0.0% | 499.8 ± 125.0 µm | 314.1 ± 120.2 µm | 100.0% | 87.5 ± 33.9 s |
| | | SG | 5 | 1 | 20.0% | 159.5 ± 37.4 µm | 126.4 ± 42.6 µm | 75.0% | 150.6 ± 28.2 s |
| | Robot Driven | TR | 9 | 2 | 22.2% | 106.8 ± 36.8 µm | 81.9 ± 40.7 µm | 71.4% | 309.8 ± 174.6 s |
| | | AR | 6 | 0 | 0.0% | 103.3 ± 4.5 µm | 75.2 ± 16.4 µm | 83.3% | 165.1 ± 59.2 s |